\documentclass[journal]{IEEEtran}

\ifCLASSINFOpdf
\else
   \usepackage[dvips]{graphicx}
\fi
\usepackage{url}
\usepackage{graphicx}
\usepackage{amsfonts}
\usepackage{booktabs, multirow, placeins, float, indentfirst, mathrsfs, tabularx, adjustbox}
\usepackage{array}
\usepackage{pifont}
\usepackage{cancel}
\usepackage{titlesec}
\usepackage{hyperref}
\usepackage{amsmath}
\usepackage{amssymb}
\titlespacing{\section}{0pt}{1em}{0.5em}
\titlespacing{\subsection}{0pt}{0.9em}{0.3em}

\begin{document}

\title{GPTSee: Enhancing Moment Retrieval and Highlight Detection via Description-Based Similarity Features}
\author{
Yunzhuo Sun, Yifang Xu, Zien Xie, Yukun Shu, and Sidan Du, \IEEEmembership{Member, IEEE}
\thanks{Manuscript received August 10, 2023; revised November 23, 2023; accepted November 23, 2023. Date of publication December 16, 2023; date of current version December 23, 2023. The associate editor coordinating the review of this manuscript and approving it for publication was Yue Gao (Corresponding author: Sidan Du.)}
\vspace{-1mm}
\thanks{Yunzhuo Sun and Yukun Shu are with the School of Physics and Electronics, Hubei Normal University, Huangshi 435002, China (e-mail: sunyunzhuo98@outlook.com; Shuyk@hbnu.edu.cn)}
\vspace{-1mm}
\thanks{Yifang Xu, Zien Xie and Sidan Du are with the School of Electronic Science and Engineering, Nanjing University, Nanjing 210000, China (e-mail: xyf@smail.nju.edu.cn; xze@smail.nju.edu.cn; coff128@nju.edu.cn).}

}

\markboth{IEEE SIGNAL PROCESSING LETTERS, VOL. 30, 2023}
{Shell \MakeLowercase{\textit{et al.}}: Bare Demo of IEEEtran.cls for IEEE Journals}
\maketitle

\begin{abstract}
Moment retrieval (MR) and highlight detection (HD) aim to identify relevant moments and highlights in video from corresponding natural language query. Large language models (LLMs) have demonstrated proficiency in various computer vision tasks. However, existing methods for MR\&HD have not yet been integrated with LLMs. In this letter, we propose a novel two-stage model that takes the output of LLMs as the input to the second-stage transformer encoder-decoder. First, MiniGPT-4 is employed to generate the detailed description of the video frame and rewrite the query statement, fed into the encoder as new features. Then, semantic similarity is computed between the generated description and the rewritten queries. Finally, continuous high-similarity video frames are converted into span anchors, serving as prior position information for the decoder. Experiments demonstrate that our approach achieves a state-of-the-art result, and by using only span anchors and similarity scores as outputs, positioning accuracy outperforms traditional methods, like Moment-DETR.

\end{abstract}


\IEEEpeerreviewmaketitle

\section{INTRODUCTION}
As the internet and video production technology evolve rapidly, users upload hundreds of millions of videos to various platforms daily. How to effectively search and browse through such a vast amount of content has attracted widespread attention. Given a video and a natural language query, video moment retrieval (MR) \cite{ExCL-2019, 2D-TAN, xu2024vtg} strives to retrieve the most relevant spans, each comprising a start and an end moment. On the other hand, video highlight detection (HD) \cite{SL-Module-2021, sLSTM-2016, } aims to predict moment-wise highlight scores across the whole video. In this letter, we focus on MR\&HD simultaneously due to their shared characteristics, notably the need to learn the similarity between the textual query and video moments.

With the recent surge in large language models (LLMs) like LLaMA \cite{LLaMA-2023} and GPT-4 \cite{ChatGPT4}, an emerging trend is to adapt these expansive models to computer vision tasks \cite{MiniGPT4-2023, Multimodal-GPT-2023, VideoChat-2023, VideoChatGPT-2023}. This transition has unveiled impressive capabilities; for instance, MiniGPT-4 \cite{VideoChatGPT-2023} can create websites from handwritten drafts and generate detailed image captions. Moreover, VideoChat \cite{VideoChat-2023} and Video-ChatGPT \cite{VideoChatGPT-2023} have demonstrated adaptability to certain video understanding subtasks such as video summarization and video question answering. However, existing GPT-based video models encounter difficulties with more fine-grained subtasks like MR\&HD. This is due to two main reasons. Firstly, MR\&HD necessitates modeling of moment-level features. However, the upper limit on the context length in large models poses a significant constraint, reducing their performance. Secondly, these large models lack dedicated modules explicitly designed for MR\&HD \cite{Survey-TSGV-2022}.

In this paper, we propose a two-stage stepwise optimization model, utilizing the output of the LLMs as input to the transformer encoder-decoder \cite{Transformer}. First, we extract a frame every two seconds from the video, converting them into textual descriptions using MiniGPT-4 \cite{MiniGPT4-2023}. Query rewrites with identical semantics are generated with the same model to explore semantic information. We then calculate the semantic similarity between the content description and queries, identifying the range for continuous video frames with high similarity, termed span anchors. Finally, the features and span anchors from MiniGPT-4 are input into the second-stage transformer's encoder and decoder, respectively. 
\begin{figure}[t!]
  \centering
  \includegraphics[width=\linewidth]{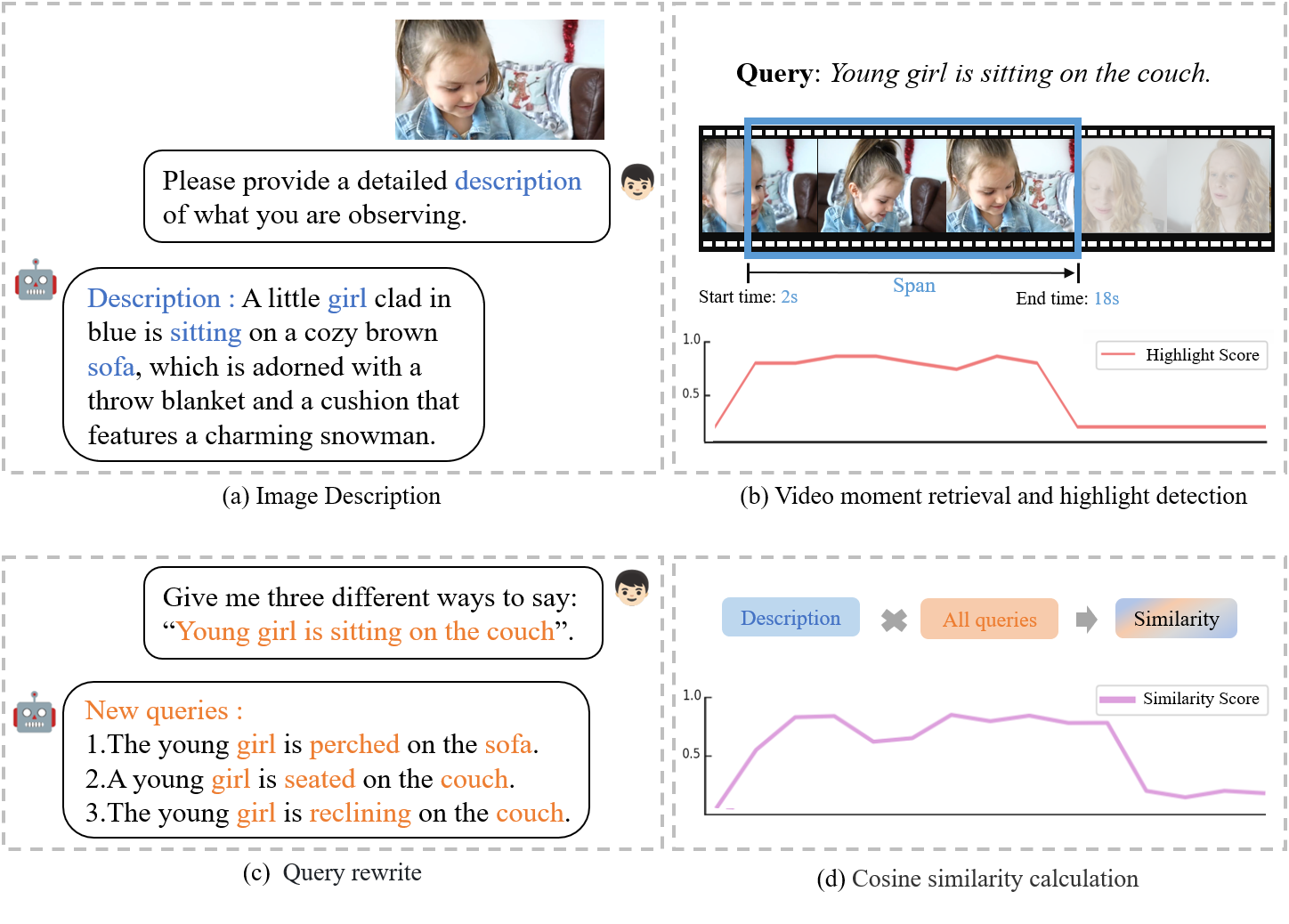}
  \caption{
    (a) Describe video frame content with GPT. (b) Examples of video moment retrieval and highlight detection (MR\&HD) tasks. (c) Rewrite queries using GPT. (d) Calculate the cosine similarity between the image description and the rewritten query.
  }
  \label{Fig:MRHD}
\end{figure}
\begin{figure*}[t!]
  \centering
  \includegraphics[width=\linewidth]{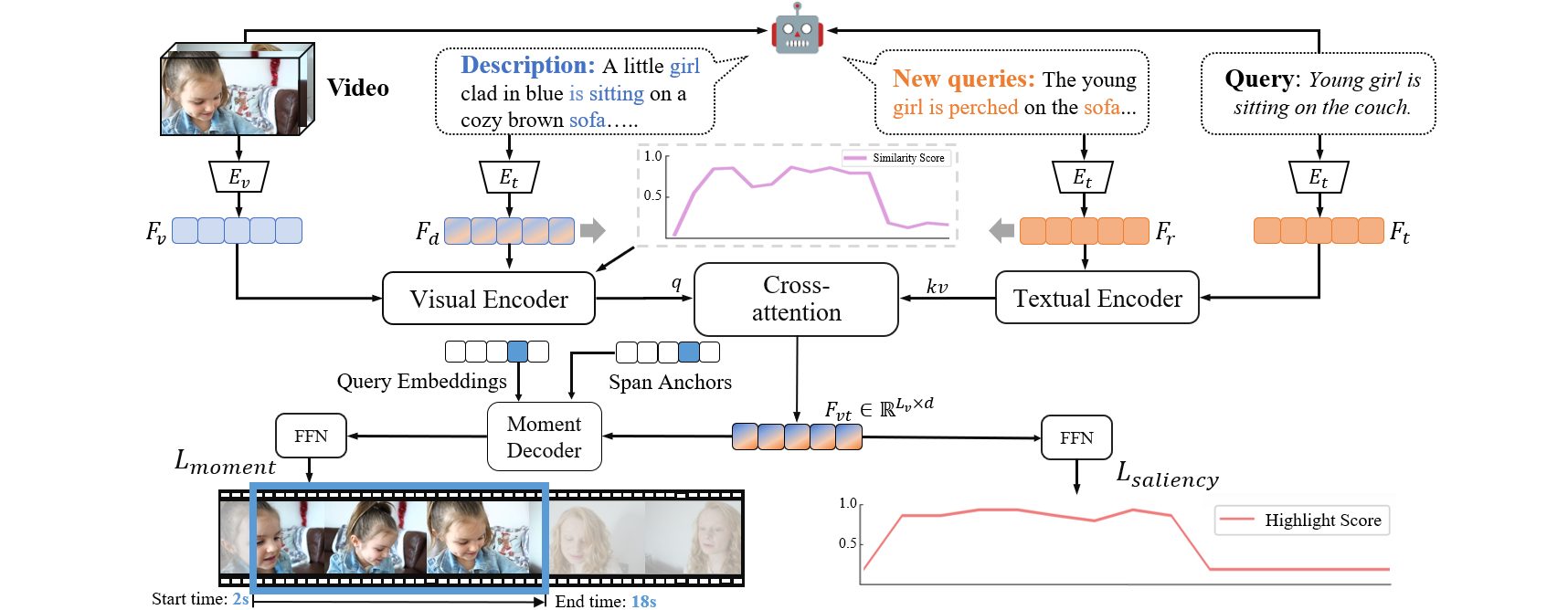}
  \caption{
    An overview of our proposed model GPTSee. Video frames and query text are initially fed into MiniGPT4, generating corresponding image content descriptions and semantically rewritten queries. Subsequently, the visual extractor ${E}_v$ and text extractor ${E}_t$ obtain features from these descriptions and rewritten queries, which are input into their respective visual and text encoders. In parallel,  similarity scores are calculated based on the semantic similarity between the image content descriptions of key video frames and rewritten queries. The visual encoder jointly receives these scores, concatenated with the image description. The encoded visual and text features interact through a cross-attention mechanism, resulting in the cross-modal features $F_{vt}$. This feature is then directly processed by an FFN to derive the highlight scores for the HD task. Frames bearing consecutive high similarity scores form a range, referred to as span anchors, serving as prior position information for the moment decoder. Subsequently, for the MR task, this decoder establishes the start and end positions of video moments.
  }
  \label{Figure2}
\end{figure*}
In fact, due to the instability of bipartite graph matching, DETR-like models tend to underperform. However, the high-quality features and span anchors facilitate the positioning in the second-stage model, thereby enhancing the final results.

Overall, our main contributions are as follows:
\begin{enumerate}
    \item We use LLMs to generate detailed descriptions of images and rewrite queries, then compute the semantic similarity scores between them. The mentioned operation introduces three novel features for the MR\&HD task.
    \item We optimized the decoder module by leveraging high-quality prior positional information from the first stage, enhancing model performance.
    \item We have conducted extensive experiments on the QVHighlights dataset, demonstrating that our method performs better than the current state-of-the-art approaches
\end{enumerate}

\section{METHOD}
\subsection{Overview}

Given an untrimmed video \( V \in \mathbb{R}^{N_v \times H \times W \times 3} \) containing \( N_v \) moments and a natural language query \( T \in \mathbb{R}^{N_t} \) with \( N_t \) words, the task of MR\&HD aims to localize all boundaries  \( B \in \mathbb{R}^{N_b \times 2} \), each comprising a start moment and an end moment, that are highly relevant to \( T \). Simultaneously, it predicts moment-wise highlight scores \( H \in \mathbb{R}^{N_v} \) for the entire video. The overall structure of our approach, designed based on the foundational principles of Moment-DETR \cite{MomentDETR-2021}, is depicted in Figure \ref{Figure2}.

Our process begins with generating detailed image descriptions and query rewrites. Utilizing MiniGPT-4 \cite{MiniGPT4-2023}, we produce natural language descriptions for each video frame and create query rewrites that retain semantic similarity while introducing syntactic variations. With the aid of CLIP \cite{CLIP-2021}, visual features ${F}_v \in \mathbb{R}^{L_v \times d_v}$ and textual features ${F}_t \in \mathbb{R}^{L_t \times d_t}$ are extracted from raw videos and queries, respectively. Similarly, the CLIP text encoder ${E}_t$ extracts features from the descriptions ${F}_d$ and query rewrites ${F}_r$. Corresponding encoders process these features to produce visual tokens $\tilde{F}_v \in \mathbb{R}^{N_v \times d}$ and textual tokens $\tilde{F}_t \in \mathbb{R}^{N_t \times d}$, which are then fused by a cross-modal interaction module. The computation of similarity scores \( S \in \mathbb{R}^{N_s \times 1} \) between the visual and textual features identifies span anchors ${A} \in \mathbb{R}^{L_v \times 2}$, used as prior positional information in the decoder. The final stage employs a linear layer and sigmoid activation to estimate moments and highlight scores in the prediction head.

\subsection{Image Detail Description}

Videos contain rich semantic information. Previous work has predominantly focused on extracting features from specific aspects of images, such as optical flow \cite{RAFT-2022, GMFlow-2022}, depth maps \cite{Xu2021DAFFN, Xu2021PFAN}, achieving noteworthy results. However, these prior approaches have overlooked the potential of translating the visual content of images into natural language descriptions. Inspired by the way humans perceive content within videos, our approach translates visual content into comprehensive text descriptions and feeds them into the encoder as an innovative feature form. This method encodes video content from a textual standpoint, introducing a new dimension to the field.

We employ MiniGPT-4 \cite{MiniGPT4-2023} to generate natural language descriptions for each frame of the input videos, as shown in Figure. \ref{Fig:MRHD} (a). MiniGPT-4, which integrates advanced language models with visual perception components, can generate content-rich and semantically coherent text descriptions. For instance, given a video frame depicting a little girl sitting on a sofa, MiniGPT-4 might produce a description such as "A little girl clad in blue is sitting on a cozy brown sofa." These descriptions serve as additional contextual information, feeding into the model to enrich its comprehension of the video content.
\subsection{Query Rewriting}
To fully exploit the semantic information of queries, we have designed a query rewriting module, primarily relying on two strategies: semantic equivalent transformation and synonym substitution.    For example, given an original query, "Young girl is sitting on the couch" can be transformed into "The young girl is perched on the sofa" through semantic equivalent transformation or into "A young girl is seated on the couch" by applying synonym substitution.

We guide MiniGPT-4 \cite{MiniGPT4-2023} to generate syntactically distinct sentences but semantically close sentences. In this way, the model can interpret and understand queries from various perspectives, thus enhancing its ability to handle ambiguous or unclear queries. The quality of these rewritten queries is validated by computing their semantic similarity to the original queries.  
\begin{align}
M(q, q_0) = \min \left( 1, \frac{P(q_0|q)}{P(q|q)} \right)
\end{align}
where \( M(q, q_0) \) is the measure of semantic similarity between the original query \( q \) and its paraphrase \( q_0 \), \( P(q_0|q) \) is the probability of a paraphrase \( q_0 \) given the original question \( q \), and \( P(q|q) \) is used to normalize different distributions.

\begin{figure}[t!]
  \centering
  \includegraphics[width=\linewidth]{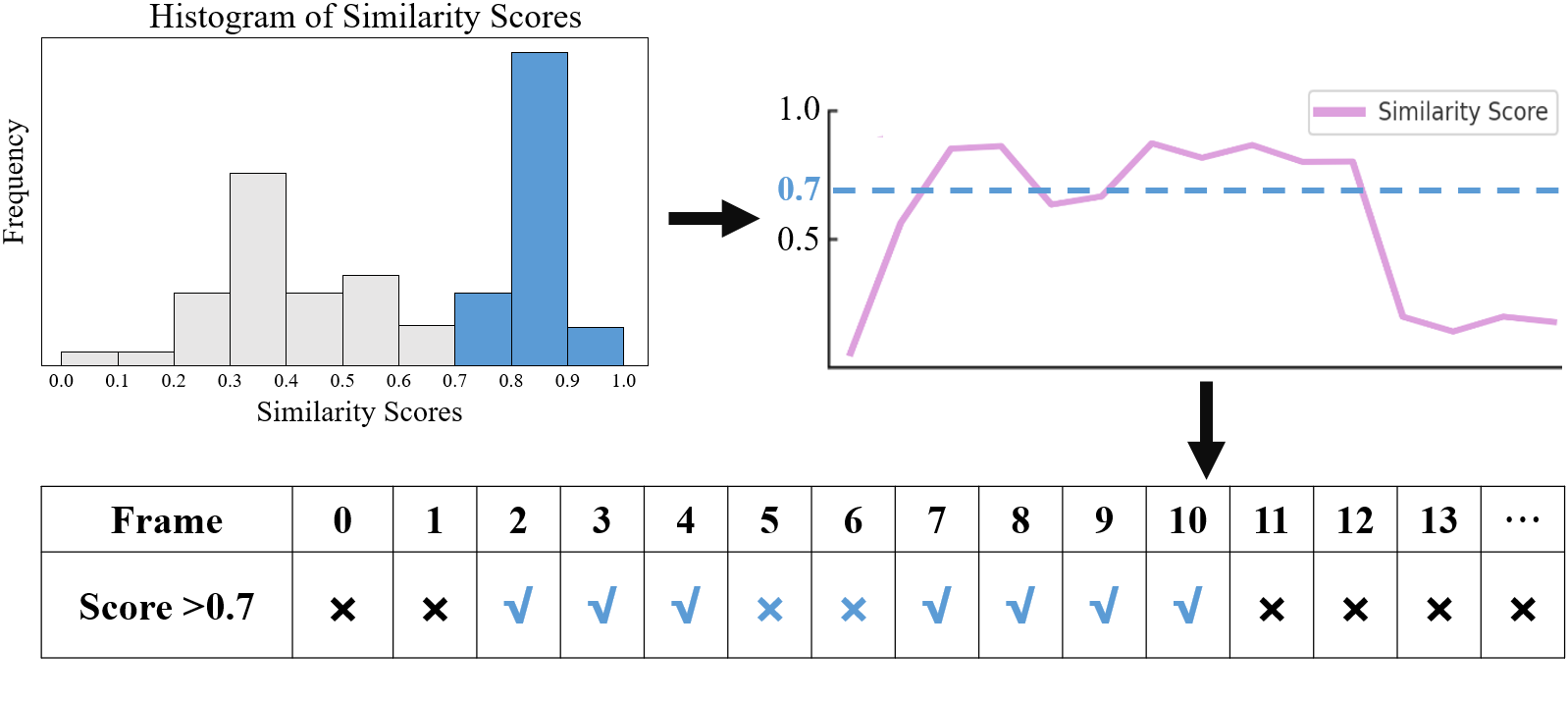}
  \caption{
     Determine the threshold and then collect the indices of all frames whose similarity scores exceed this threshold.
  }
  \label{Fig:3}
\end{figure}
\subsection{Similarity Calculation and Span Anchors }

When humans perform MR\&HD tasks, it is natural to compare semantic similarity between video content and query text to determine relevance \cite{Zhen2019DeepSCMRetrieval, Chun2021ProbabilisticEC}. Inspired by this, we analyze the similarity \( S \in \mathbb{R}^{N_s \times 1} \) and set thresholds to identify the relevant range. As shown in Figure \ref{Fig:3}, we calculate the cosine similarity between the image description and the rewritten query to quantify their relevance. Due to the inconsistent distribution of similarity scores among video groups, employing a fixed threshold for determining relevance is not advisable. Therefore, we analyze the distribution in each group, setting the threshold as the third most common value. Scores above this threshold are marked with a $\checkmark$, while those below are marked with a $\times$. If the number of $\times$ marks is less than six, the range is considered non-relevant and is defined as span anchors ${A} \in \mathbb{R}^{L_v \times 2}$. Experimental results demonstrate that, without additional training and solely utilizing the span anchors and similarity score, our model can surpass Moment-DETR \cite{MomentDETR-2021} on the MR\&HD task. According to \cite{DAB-DETR-2022}, we feed ${S}$ into the visual encoder and process ${A}$ through a feed-forward network (FFN), incorporating it into query embeddings and then providing it into the moment decoder.

\subsection{Moment Decoder and Prediction Heads}

We integrate visual tokens and textual tokens via cross-attention \cite{Zhen2022DeepMTL, Cheng2022ViSTA, Xu2023QGRDSN} to form ${F}_{vt} \in \mathbb{R}^{L_v \times d}$. An FFN with ReLU \cite{ReLU-2013} predicts normalized moment center and width. Class label prediction utilizes a softmax linear layer. Predicted moments are assigned foreground or background based on alignment with ground truth. Another linear layer predicts highlight scores, as $H \in \mathbb{R}^{N_v}$. 

\textit{Moment Retrieval Loss}. \( L_m \), measuring the  between predicted \( \hat{m} \) and ground-truth moments \( m \), is defined as:
\begin{align}
L_m &= \lambda_{L1} ||m - \hat{m}||_1  + \lambda_{iou} L_{iou}(m, \hat{m})
\end{align}
where \( \lambda_{L1} \) and \( \lambda_{iou} \) are real-valued hyperparameters. It combines the L1 loss and the generalized Intersection-over-Union (IoU) loss \( L_{iou} \) \cite{GIoU-2019},  which computes the temporal overlap between \( \hat{m} \) and \( m \) .

\textit{Cross-entropy Loss}. We utilize the weighted binary cross-entropy loss to categorize predicted spans into foreground or background.  This can be mathematically represented as:
\begin{equation}
L_{\text{cls}} = -\sum_{i=1}^{L_s} \left[ w_p z_i \log(p_i) + (1-z_i) \log(1-p_i) \right]
\end{equation} 
In this equation, $p_i$ and $z_i$ denote the forecasted probability of the foreground and its respective label.  The foreground label is attributed with a higher weight $w_p$ to alleviate label imbalance.

\textit{Highlight Detection Loss}
The Highlight Detection Loss is designed to optimize the highlight score for each moment. This loss is computed using hinge loss across two distinct sets of segments: 
\begin{align}
L_{\text{h}} &= \max(0, \delta + H(t_{\text{low}}) - H(t_{\text{high}})) \nonumber \\
&\quad + \max(0, \delta + H(t_{\text{out}}) - H(t_{\text{in}}))
\end{align}
Here, the first set comprises a high-score segment (\( t_{\text{high}} \)) and a low-score segment (\( t_{\text{low}} \)) within the actual temporal moments. The second set includes one segment (\( t_{\text{in}} \)) inside and another (\( t_{\text{out}} \)) outside the actual temporal moments.

\textit{Total Loss}. The total loss is computed as a linear combination of the above losses:
\begin{align}
L_{\text{total}} &= L_{\text{m}} + \lambda_{\text{cls}}L_{\text{cls}} + \lambda_{\text{h}}L_{\text{h}}
\end{align}

\section{Experiments}

\subsection{Evaluation Dataset and Metrics Selection}

Based on the QVHighlights \cite{MomentDETR-2021, Xu2023MHDETR} dataset, we evaluated our model. QVHighlights is the only publicly accessible dataset with ground-truth labels for MR\&HD.  The dataset comprises 10,148 YouTube videos. Each video in the dataset is annotated with an unstructured textual query, associated time spans, and scores for significant moments. We followed a widely accepted data partitioning scheme (training, validation, testing) utilized in recent studies to ensure a fair comparison.

To measure the effectiveness of moment retrieval (MR), we employed metrics like Recall@1 with thresholds of 0.5 and 0.7, mean average precision (mAP) with intersection over union (IoU) thresholds of 0.5 and 0.75, and consolidation of mAP at various IoU thresholds in the range of 0.5 to 0.95 incremented by 0.05 were used. For highlight detection (HD), mAP along with HIT@1 was employed, wherein HIT@1 accounts for instances where the moment with the highest score is correctly identified.

\vspace{-2mm}
\subsection{Details of Implementation}

Our model integrates a visual encoder, a textual encoder, and a cross-modal interaction module, all equipped with a singular attention layer, allowing seamless communication between the visual and textual information. Additionally, our moment decoder includes four self-attention layers. We apply a dropout rate 0.1, enhanced by post-normalization style layer normalization \cite{LayerNorm-2016} and ReLU \cite{ReLU-2013} activation functions.The loss function's hyperparameters are set as follows: $\lambda_{L1} = 10$, $\lambda_{iou} = 1$, $\lambda_{cls} = 4$, $\lambda_{h} = 1$ and $w_{p} = 10$. We use the AdamW optimizer \cite{AdamW-2017}, with a learning rate set at 2e-4 and a weight decay parameter of 1e-4. We conducted training over 200 epochs with a batch size of 32, utilizing 8 RTX 3090.

\subsection{Experimental Results}

Initially, we provide an extensive comparison of our proposed GPTSee model with preceding state-of-the-art models on the QVHighlights dataset, as presented in Table \ref{tab:QVHighlights-test}. Across all metrics, our model consistently excels over the UniVTG \cite{UniVTG-2023}. Notably, there are increments of 5.86\% and 1.03\% in MR-mAP Avg. and HD-mAP, respectively, underscoring our model's robustness and efficacy in the MR\&HD task.

Following that, we perform a targeted comparison between our GPTSee model and Moment-DETR \cite{MomentDETR-2021} in Table \ref{tab:QVHighlights-test-only GPT}, demonstrating the benefits of employing span anchors for evaluation. GPTSee surpasses Moment-DETR, even when using only span anchors ($A$) and similarity ($S$) as output, particularly for MR-mAP, HD-mAP, and HD-HIT@1 metrics.

\subsection{Ablation Studies}

Ablations on Transfer Capability: As shown in Table \ref{tab:QVHighlights-test-only GPT}, the integration of $A$ and $S$ into Moment-DETR and UMT was tested to validate the generality of our two-stage approach. The results reveal that this method improves accuracy. Notably, using $A + S$ alone for final localization, the HD-HIT@1 scores excel, showing strong performance in local highlight detection but only average overall. By incorporating the similarity into the encoder, our method achieved a significant enhancement in overall HD tasks, along with a slight improvement in local.

\textit{Ablations on LLMs}: As depicted in Table \ref{tab:query_analysis}, when selecting models for generating image detail descriptions and rewriting queries, a comparison was made among VideoChat \cite{VideoChat-2023}, Video-ChatGPT \cite{VideoChatGPT-2023}, and MiniGPT-4, utilizing the same prompt for generating descriptions and rewriting queries. The conclusion ascertains that MiniGPT-4 slightly outperforms the other two. This is likely due to MiniGPT-4 generating fewer irrelevant words during sentence creation, thus having an advantage in semantic similarity computation.

\textit{Ablations on Module Configuration}: As shown in Table \ref{tab:ablation-features}, to validate the effectiveness of each model component, several baseline models were constructed with varying components. The analysis reveals that the description and span anchors contribute most significantly to the overall model performance, while query rewrite provides only a marginal contribution.

\begin{table}[t!]
\caption{Performance comparison on QVHighlights test split.}
\label{tab:QVHighlights-test}
\centering
\begin{adjustbox}{max width=0.9\linewidth}
\begin{tabular}{@{}cccccccc@{}}
\toprule
\multicolumn{1}{c}{\multirow{3}{*}{\textbf{Methods}}} & \multicolumn{5}{c}{\textbf{MR}} & \multicolumn{2}{c}{\textbf{HD}} \\
\multicolumn{1}{c}{} & \multicolumn{2}{c}{R1} & \multicolumn{3}{c}{mAP} & \multicolumn{2}{c}{$\geq$ Very Good} \\
\cmidrule(lr){2-3}
\cmidrule(lr){4-6}
\cmidrule(lr){7-8}
\multicolumn{1}{c}{} & @0.5 & @0.7 & @0.5 & @0.75 & Avg. & mAP & HIT@1 \\
\midrule
CAL \cite{CAL-2019} & 25.49 & 11.54 & 23.40 & 7.65 & 9.89 & - & - \\
XML \cite{TVR-dataset-2020} & 41.83 & 30.35 & 44.63 & 31.73 & 32.14 & 34.49 & 55.25 \\
XML+ \cite{TVR-dataset-2020} & 46.69 & 33.46 & 47.89 & 34.67 & 34.90 & 35.38 & 55.06 \\ \midrule
Moment-DETR \cite{MomentDETR-2021} & 52.89 & 33.02 & 54.82 & 29.40 & 30.73 & 35.69 & 55.60 \\
UMT \cite{UMT-2022} & 56.23 & 41.18 & 53.83 & 37.01 & 36.12 & 38.18 & 59.99 \\
UniVTG \cite{UniVTG-2023} & 58.86 & 40.86 & 57.60 & 35.59 & 35.47 & 38.20 &  60.96 \\
GPTSee (Ours) & \textbf{62.84} & \textbf{48.01} & \textbf{61.92} & \textbf{42.55} & \textbf{41.33} & \textbf{39.23} & \textbf{62.80} \\ 
\bottomrule
\end{tabular}
\end{adjustbox}
\end{table}


\begin{table}[t!]
\caption{Performance comparison of span anchors and similarity scores as direct outputs on QVHighlights test split.}
\label{tab:QVHighlights-test-only GPT}
\centering
\begin{adjustbox}{max width=0.9\linewidth}
\begin{tabular}{@{}cccccc@{}}
\toprule
\multirow{2}{*}{\textbf{Model.}} & \multicolumn{3}{c}{\textbf{MR}} & \multicolumn{2}{c}{\textbf{HD} ($\geq$VG)} \\
\cmidrule(lr){2-4}
\cmidrule(lr){5-6}
 & R1@0.5 & R1@0.7 & mAP Avg. & mAP & HIT@1 \\ 
\midrule
A + S  & 56.55 & 31.60 & 32.27 & 36.11 & \textbf{61.65} \\ 
\midrule
Moment-DETR \cite{MomentDETR-2021}  & 54.82 & 29.40 & 30.73 & 35.69 & 55.60 \\
Moment-DETR + A + S & \textbf{60.20} & \textbf{34.88} & \textbf{36.29} & \textbf{36.45} & 59.04 \\ 
\midrule
UMT \cite{UMT-2022}  & 53.83 & 37.01 & 36.12 & 38.18 & 59.99 \\
UMT \cite{UMT-2022} + A + S & \textbf{60.74} & \textbf{40.05} & \textbf{38.01} & \textbf{38.58} & 61.07 \\
\bottomrule
\end{tabular}
\end{adjustbox}
\end{table}

\begin{table}[t!]
\caption{COMPARISON OF SPAN ANCHORS AND SIMILARITY SCORES FROM DIFFERENT LLMS IN QVHIGHLIGHTS TEST SPLIT.}
\label{tab:query_analysis}
\centering
\begin{adjustbox}{max width=0.9\linewidth}
\begin{tabular}{@{}cccccc@{}}
\toprule
\multirow{2}{*}{\textbf{Model.}} & \multicolumn{3}{c}{\textbf{MR}} & \multicolumn{2}{c}{\textbf{HD} ($\geq$VG)} \\
\cmidrule(lr){2-4}
\cmidrule(lr){5-6}
 & R1@0.5 & R1@0.7 & mAP Avg. & mAP & HIT@1 \\ 
\midrule
VideoChat\cite{VideoChat-2023} & 60.23 & 47.34 & 39.10 & 37.64 & 60.14 \\
Video-ChatGPT\cite{VideoChatGPT-2023} & \textbf{63.20} & 47.89 & 41.12 & 38.78 & 62.00 \\
MiniGPT-4\cite{MiniGPT4-2023}  & 62.84 & \textbf{48.01} & \textbf{41.33} & \textbf{39.23} & \textbf{62.80} \\ 

\bottomrule
\end{tabular}
\end{adjustbox}
\end{table}

\begin{table}[t!]
\caption{Ablation study of different features on QVHighlights test split.}
\label{tab:ablation-features}
\centering
\begin{adjustbox}{max width=0.9\linewidth}
\begin{tabular}{@{}ccccccccc@{}}
\toprule
\multicolumn{4}{c}{Features} & \multicolumn{1}{c}{MR} & \multicolumn{1}{c}{HD ($\geq$VG)} \\
\cmidrule(lr){1-4}
\cmidrule(lr){5-5}
\cmidrule(lr){6-6}
Description &  Rewritten Queries & Similarity & Span Anchors & mAP Avg. & mAP \\
\midrule
\checkmark &  &  &  & 36.54 & 36.71\\
 & \checkmark &  &  & 34.40 & 36.40 \\
 &  & \checkmark &  & 36.15 & 37.28\\
&  &  &  \checkmark & 37.52 & 37.18 \\
\midrule
\checkmark & \checkmark & & &  37.56 & 36.99\\
\checkmark & & \checkmark & &  38.36 & 37.74\\
\checkmark & & & \checkmark &  39.30 & 37.66\\
& \checkmark & \checkmark & & 37.04 & 37.44\\
& \checkmark& & \checkmark & 37.00 & 37.53\\ 
& & \checkmark & \checkmark & 38.09 & 38.10\\
\midrule
\checkmark & \checkmark & \checkmark & & 39.65 & 38.01 \\
\checkmark & \checkmark &  & \checkmark & 38.76 & 37.98 \\
\checkmark &  & \checkmark & \checkmark &40.87 & 38.61 \\
& \checkmark & \checkmark & \checkmark & 38.32 & 38.50\\
\midrule
\checkmark & \checkmark & \checkmark & \checkmark & \textbf{41.33} & \textbf{39.23} \\ 
\bottomrule
\end{tabular}
\end{adjustbox}
\end{table}

\section{Conclusion}

In this letter, we introduce an innovative two-stage model, GPTSee, which integrates LLMs' output to assist a transformer encoder-decoder architecture. The effectiveness of our model is significantly enhanced by incorporating image descriptions and rewritten queries as novel inputs and using span anchors as a priori positional information. This framework amplifies video frames and query text data utilization, eliminating the need for intricate feature extraction or elaborate training schemas characteristic of earlier approaches. Experiments on the QVHighlights dataset substantiate our model's superiority and efficacy. Future work may focus on designing more powerful LLMs or improving highlight score calculation.

\clearpage

\bibliographystyle{IEEEtran}
\bibliography{paper.bib}

\end{document}